\documentclass[letterpaper, 10 pt, journal, twoside]{IEEEtran}

\usepackage{color}
\usepackage{soul}
\sethlcolor{yellow}

\usepackage{amsmath}
\usepackage{amssymb}
\usepackage{amsfonts}
\usepackage{amsthm}
\usepackage{mathrsfs}

\usepackage[utf8]{inputenc}
\usepackage[T1]{fontenc}
\usepackage{times}
\usepackage{textcomp}

\usepackage{graphicx}
\usepackage{array}
\usepackage{booktabs}
\usepackage{tabularx}
\usepackage{makecell}
\usepackage{colortbl}
\usepackage{tablefootnote}
\usepackage{multirow}
\usepackage{multicol}
\usepackage[caption=false,font=normalsize,labelfont=sf,textfont=sf]{subfig}
\usepackage{stfloats}
\usepackage{placeins}
\usepackage{xcolor}
\usepackage{enumerate}
\usepackage{paralist}
\usepackage{comment}
\usepackage{verbatim}

\usepackage{algorithm}
\usepackage{algpseudocode}

\usepackage[numbers]{natbib}

\usepackage{hyperref}
\hypersetup{
    colorlinks=true,
    linkcolor=blue,
    filecolor=magenta,
    urlcolor=cyan
}

\theoremstyle{definition}

\definecolor{Diff-ILC}{HTML}{0053D6}
\newcommand{\DiffILC}{\textbf{\text{\textcolor{Diff-ILC}{Diff-ILC}}}}

\algrenewcommand\algorithmicrequire{\textbf{Input:}}
\algrenewcommand\algorithmicensure{\textbf{Return:}}
\begin{document}

\algrenewcommand\algorithmicindent{0pt}
\title{Differentiable Iterative Learning Control for Black-Box Position-Commanded Manipulators}

\author{Mohammed Al-Ghaili, Yakoop Qasim, and Hassan Almubarak
% \thanks{$^{\ast}$\scriptsize Corresponding author.\\ Control and Instrumentation Engineering Department, KFUPM, Dhahran, Saudi Arabia.}%   
%\thanks{$^{\ast}$ Corresponding author.}%   
\thanks{M. Al-Ghaili, Y. Qasim, and H. Almubarak are with the Control \& Instrumentation Engineering Department, KFUPM, Dhahran, Saudi Arabia. Email: {\tt \{g202424100, g202423660, halmubarak\}@kfupm.edu.sa}}%
\thanks{Project website: \url{https://cortx-lab.github.io/projects/differentiable-ilc/}.}
}

%\author{IEEE Publication Technology,~\IEEEmembership{Staff,~IEEE,}
        % <-this % stops a space
%\thanks{This paper was produced by the IEEE Publication Technology Group. They are in Piscataway, NJ.}% <-this % stops a space
%\thanks{Manuscript received April 19, 2021; revised August 16, 2021.}}

% The paper headers
%\markboth{Journal of \LaTeX\ Class Files,~Vol.~14, No.~8, August~2021}%
%{Shell \MakeLowercase{\textit{et al.}}: A Sample Article Using IEEEtran.cls for IEEE Journals}

%\markboth{IEEE Robotics and Automation Letters. Preprint Version. Accepted December, 2021}
%{Almubarak \MakeLowercase{\textit{et al.}}: Safety Embedded Differential Dynamic Programming} 

% \IEEEpubid{0000--0000/00\$00.00~\copyright~2021 IEEE}
% Remember, if you use this you must call \IEEEpubidadjcol in the second
% column for its text to clear the IEEEpubid mark.

\maketitle

\begin{abstract}
High-accuracy repetitive tracking is challenging for position-commanded manipulators whose low-level controllers are inaccessible, while constrained model-based iterative learning control (ILC) can require substantial manual parameter tuning. We present Differentiable Iterative Learning Control (\DiffILC{}), an online bilevel framework that adapts the parameters of a constrained ILC update between physical robot executions. A closed-loop differential model is identified once from paired smooth perturbations and subsequently kept fixed. At each trial, a constrained quadratic program computes a feasible command correction, while a real-anchored next-trial prediction defines a task-level objective. Hypergradients are obtained by implicitly differentiating the locally active KKT system, enabling multiple computational parameter updates without additional robot executions. Experiments on a 6-DoF DOBOT CR3 across unseen planar and three-dimensional trajectories show substantially faster physical-trial convergence than fixed-parameter model-based ILC. Across the four primary trajectories, after only two trial-to-trial updates, \DiffILC{} achieves $78.8\%$ and $76.4\%$ lower mean joint-vector and Cartesian RMSE, respectively, than fixed-parameter model-based ILC.

\end{abstract}

%\begin{IEEEkeywords}
%Optimization and Optimal Control, Robot Safety, Constrained Motion Planning.
%\end{IEEEkeywords}
\vspace{-4mm}
\section{Introduction}
\label{sec:introduction}

Iterative learning control (ILC) is well suited to finite-duration repetitive
tasks because it improves the command applied on subsequent trials using
tracking information from previous executions
\cite{saab2022,liu2021,chen2021,ta2023}. This is particularly attractive for
commercial robotic manipulators, which often expose joint-position command
interfaces while keeping their embedded controllers and actuator-level
dynamics inaccessible. In this setting, tracking performance can be improved
between repetitions without modifying the internal feedback controller or
requiring torque-level robot models \cite{lee2021industrial}.

Optimization-based ILC computes the trial-to-trial correction over the complete
trajectory horizon and can directly incorporate tracking objectives,
regularization, and motion constraints. Constrained ILC has been formulated as
a finite-dimensional quadratic program \cite{mishra2011constrained}, while
data-driven norm-optimal formulations have incorporated experimentally
estimated input-output behavior together with actuator limits
\cite{janssens2013datadriven}. Such methods can provide substantial tracking
improvement, but their performance depends strongly on parameters that balance
tracking-error reduction, correction magnitude, temporal smoothness, and
admissible trial-to-trial modification.

Reducing this tuning burden has therefore received considerable attention.
Parameter optimization, adaptive learning, and data-driven model updates
have been studied in ILC
\cite{owens2003,saab2022,liu2021,chen2021}, while more recent work has explored
self-parametrization for data-driven ILC \cite{meindl2025iterative} and
evolutionary gain adaptation for industrial robot tracking
\cite{wang2026dego}. However, parameter adaptation is more challenging in
constrained optimization-based ILC because the parameters affect the solution
of the finite-horizon optimization problem itself. Regularization and
smoothness weights modify the objective, while a correction bound can alter
the active constraint set. Task-level adaptation therefore requires
sensitivities through the constrained optimal solution.

\begin{figure}[t]
    \centering
    \includegraphics[width=\linewidth]{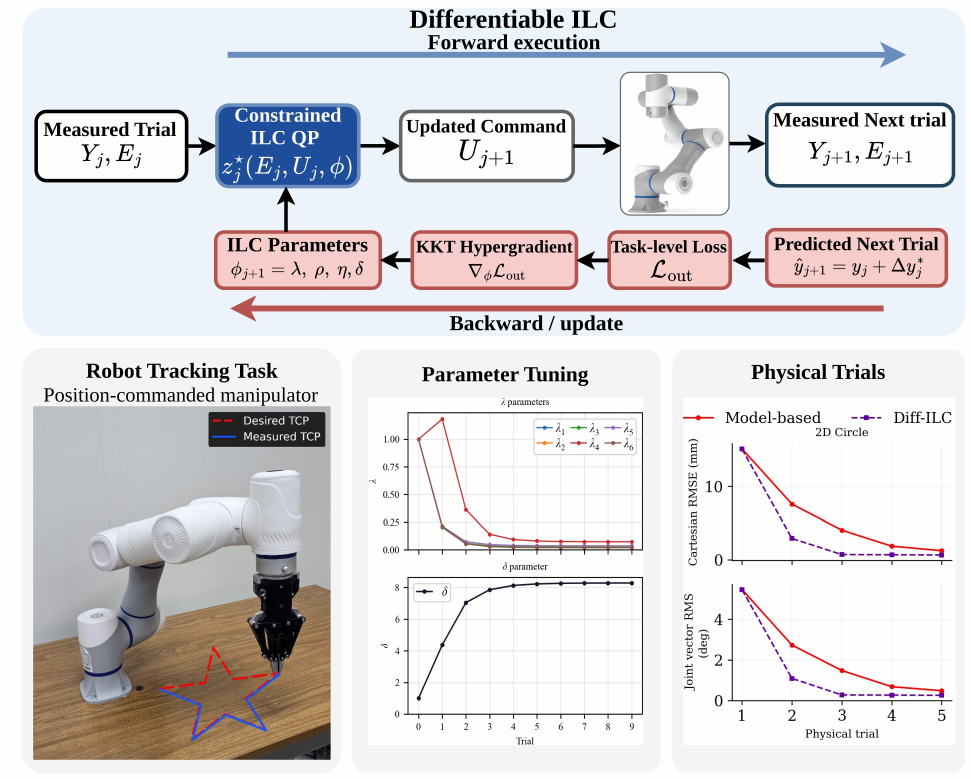}
    \caption{\DiffILC{} combines constrained model-based ILC with implicit
    differentiation through the locally active KKT system to adapt ILC
    parameters between physical robot executions. This online adaptation
    accelerates tracking-error reduction over the first few repetitions on the
    DOBOT CR3.}
    \label{fig:graphical-abstract}
    \vspace{-6mm}
\end{figure}

Differentiable optimization provides a systematic mechanism for obtaining such
sensitivities. OptNet and differentiable convex optimization layers established
general approaches for differentiating through parameterized constrained
optimization problems \cite{amos2017,agrawal2019differentiable}, while
differentiable MPC propagated gradients through optimal-control solutions
\cite{amos2018differentiable}. More recent methods such as DiffTune,
DiffTune-MPC, and differentiable robust MPC use these sensitivities for
controller adaptation
\cite{cheng2024difftune,tao2024difftune,oshin2024differentiablerobustmpc}.
The repetitive-learning setting considered here introduces an additional
challenge: each new plant evaluation requires execution of the complete
physical trajectory, the closed-loop robot dynamics are unavailable, and the
parameters to be adapted enter both the objective and active constraints of
the ILC problem.

To address these challenges, we introduce \textbf{Differentiable Iterative
Learning Control (\DiffILC{})}, an online bilevel framework for constrained
model-based ILC on black-box position-commanded manipulators. As illustrated
in \autoref{fig:graphical-abstract}, a closed-loop data-driven differential model is
developed once from paired smooth perturbations and subsequently kept fixed.
At each physical trial, a constrained quadratic program computes a feasible
command correction from the latest measured tracking error. A real-anchored
next-trial prediction then evaluates the resulting joint- and Cartesian-space
tracking performance together with command smoothness. The task-level
objective is differentiated through the locally active
Karush--Kuhn--Tucker (KKT) system of the inner ILC problem, allowing multiple
computational updates of the joint-specific regularization and smoothness
parameters and the shared correction bound between physical robot
executions.

The main contributions of this work are:
\begin{itemize}
    \item We develop a data-driven closed-loop differential model for
    command-level constrained ILC on position-commanded manipulators with
    inaccessible internal servo dynamics. The model is identified once from
    paired physical perturbations and used across subsequent trajectory geometries and execution conditions.

    \item We formulate the constrained ILC update as a differentiable
    optimization layer and introduce a bilevel adaptation scheme for
    joint-specific correction, temporal-variation, and curvature penalties
    together with a shared trial-to-trial correction bound. A real-anchored
    task-level objective is differentiated through the locally active KKT
    system using an adjoint formulation, enabling multiple parameter updates
    without additional physical executions.

    \item We validate \DiffILC{} experimentally on a 6-DoF DOBOT CR3 across
    unseen planar and three-dimensional trajectories and an additional faster
    execution condition. Using the same identified differential model,
    tracking metric, initial command, and physical motion constraints,
    \DiffILC{} achieves substantially faster tracking-error reduction over the
    first few physical repetitions than fixed-parameter constrained
    model-based ILC.
\end{itemize}

\vspace{-3mm}
\section{Problem Formulation}
\label{sec:problem}

\subsection{Repeated Trajectory Tracking}

Consider an $n_q$-joint manipulator that repeatedly executes a finite-duration trajectory over the discrete-time horizon $k\in\{0,\ldots,N-1\}$. The robot exposes a joint-position command interface while its internal controllers and low-level dynamics are unavailable to the user, which is a common setting for industrial manipulators. We therefore treat the complete closed-loop robot as an unknown trajectory-to-trajectory map
\begin{equation}
    Y_j = \mathcal{P}(U_j)
    \label{eq:robot-closed-loop-map}
\end{equation}
where $\mathcal{P}: \mathbb{R}^{N\times n_q} \rightarrow \mathbb{R}^{N\times n_q}$ denotes the unknown closed-loop mapping, and $U_j,Y_j\in\mathbb{R}^{N\times n_q}$ denote the commanded and measured joint trajectories at trial $j$, respectively.

Let $R\in\mathbb{R}^{N\times n_q}$ denote the desired joint trajectory. After trial $j$, the measured tracking error is defined as
\begin{equation}
    E_j = R-Y_j
    \label{eq:tracking-error}
\end{equation}

Iterative learning control (ILC) modifies the feedforward command between repetitions according to
\begin{equation}
    U_{j+1}=U_j+\Delta U_j
    \label{eq:general-ilc-update}
\end{equation}
where $\Delta U_j$ is a finite-horizon command correction computed from the information available after trial $j$. Unlike torque-controlled formulations, no model of the manipulator rigid-body dynamics or internal servo controller is assumed to be available. The learning controller instead relies on commanded trajectories, measured robot motion, and a data-driven approximation of the local closed-loop command-to-output response.

Each updated command must remain executable by the position-controlled robot. Let $D_1$ and $D_2$ denote first- and second-order finite-difference operators over the trajectory horizon. We require
\begin{align}
    q_{\min} &\leq U_j+\Delta U_j \leq q_{\max}
    \label{eq:problem-position-constraint}\\
    -h v_{\max} &\leq D_1(U_j+\Delta U_j) \leq h v_{\max}
    \label{eq:problem-velocity-constraint}\\
    -h^2 a_{\max} &\leq D_2(U_j+\Delta U_j) \leq h^2 a_{\max}
    \label{eq:problem-acceleration-constraint}
\end{align}
where $h$ is the sampling period, and the bounds are applied joint-wise. A trial-to-trial correction bound may additionally be imposed as
\begin{equation}
    -\delta \leq \Delta U_j \leq \delta
    \label{eq:problem-correction-bound}
 \vspace{-3mm}
 \end{equation}

\subsection{Learning Control Objective}

We consider a family of repetitive finite-duration motion tasks in which each task is specified by a desired trajectory and may differ in path geometry, execution speed, or operating condition. For a given task, the robot executes the same desired motion repeatedly, and the commanded trajectory may be modified only between successive executions.

The objective is to design a trial-to-trial learning law that achieves high-accuracy tracking while satisfying the command and motion constraints defined above. In particular, the learning controller should determine a correction $\Delta U_j$ after each physical execution such that the measured trajectory progressively approaches the desired trajectory while maintaining feasible and sufficiently smooth commanded motion.

We express the trial-to-trial correction through the general learning law
\begin{equation}
    \Delta U_j = \pi(E_j,U_j;\phi_j)
    \label{eq:update-policy}
\end{equation}
where $\pi(\cdot)$ denotes the ILC update policy, and $\phi_j$ collects the design parameters governing its behavior at trial $j$. These parameters determine how the learning law balances aggressive tracking-error reduction against the magnitude and smoothness of the command correction.

A central consideration is the number of physical repetitions required to achieve accurate tracking. Unlike an optimization performed entirely in simulation, every ILC iteration requires execution of the complete trajectory on the physical robot. Slow learning therefore increases experimental time, robot utilization, and the number of inaccurate motions executed before satisfactory tracking is achieved. The relevant objective is consequently not only low asymptotic tracking error but also rapid improvement within the first few physical repetitions.

The parameters $\phi_j$ that provide an effective balance between tracking improvement, command regularization, and trajectory smoothness may also depend on the trajectory geometry, execution speed, and operating condition. Manually retuning these parameters for each new repetitive motion can require additional physical experimentation and limits the practical scalability of model-based ILC.

The problem addressed in this work can therefore be stated as follows. 
\textit{Given a position-commanded manipulator with inaccessible closed-loop dynamics and a repetitive finite-horizon motion task, design an update law of the form \eqref{eq:update-policy} that generates feasible command corrections and drives the measured trajectory toward the desired trajectory in as few physical repetitions as possible. The learning law should remain effective across different trajectory geometries and execution speeds while automatically adapting the parameters $\phi_j$ governing the ILC update, rather than requiring trajectory-specific manual retuning.}

To this end, the next section develops a compact data-driven differential model of $\mathcal{P}$ that predicts how changes in the commanded trajectory affect the measured robot motion. This model is subsequently used in the constrained ILC update and its differentiable outer-level parameter adaptation.
\vspace{-3mm}
\section{Closed-Loop Differential Model Identification} \label{subsec:differential-model}

Data-driven and local identified models are used in the literature to construct enhanced ILC updates for systems with uncertain or inaccessible dynamics \cite{lee2021industrial, jiang2002iterative}. Here, we identify the closed-loop differential response required specifically by the constrained command-level ILC update.

The constrained ILC update requires a prediction of how a trial-to-trial change in the commanded trajectory affects the measured robot motion. Since the measured trajectory $Y_j$ is already available after trial $j$, an accurate prediction of the absolute closed-loop response $U_j\mapsto Y_j$ is not necessary. Instead, we seek a differential model that predicts the change in the measured trajectory resulting from a command correction
\begin{equation}
    \Delta U \mapsto \Delta Y
\end{equation}
The next-trial motion can then be predicted by anchoring the differential response at the measured trajectory from the current physical trial.

We approximate the closed-loop command-to-motion behavior by the discrete-time affine model
\begin{equation} \label{eq:affine-model}
    y_{k+1}=Ay_k+Bu_k+Cu_{k-1}+d
\end{equation}
where $u_k,y_k\in\mathbb{R}^{n_q}$ are the commanded and measured joint positions, $A,B,C\in\mathbb{R}^{n_q\times n_q}$ are full matrices and $d\in\mathbb{R}^{n_q}$ is a constant offset. The full matrices allow the model to capture cross-joint coupling, while the delayed command term allows for one-sample memory in the closed-loop command response.

Rather than identifying the affine model directly from absolute command and measurement data, we identify its differential response around a nominal trajectory $U^0$. Smooth excitation trajectories are commonly used in robot identification to obtain informative data while respecting motion constraints, with B-spline parameterizations providing a convenient means of generating smooth and bounded joint motions \cite{rackl2012robot}. More specifically, we construct smooth multi-joint perturbation directions $P^m$ using a B-spline basis and execute each direction symmetrically about the nominal trajectory. 
\begin{equation} \label{eq:paired-inputs}
    U^{+,m}=U^0+P^m,  \qquad  U^{-,m}=U^0-P^m
\end{equation}
    % $U^{+,m}=U^0+P^m,  \quad  U^{-,m}=U^0-P^m$
. From the corresponding measured trajectories $Y^{+,m}$ and $Y^{-,m}$, the differential command and response are constructed:
\begin{equation}
    \delta U^m=\frac{U^{+,m}-U^{-,m}}{2}, \qquad  \delta Y^m=\frac{Y^{+,m}-Y^{-,m}}{2}
    \label{eq:differential-data}
\end{equation}
The symmetric construction cancels the common nominal response and the affine offset $d$. Subtracting the paired instances of \eqref{eq:affine-model} therefore gives
\begin{equation}
    \delta y_{k+1}^m=A\delta y_k^m+B\delta u_k^m+C\delta u_{k-1}^m
    \label{eq:differential-model}
\end{equation}
which directly models the local command-to-motion response required by the ILC update.

For each perturbation direction and time sample, define
\begin{equation}
    \xi_k^m=
    \begin{bmatrix}
        (\delta y_k^m)^\top &
        (\delta u_k^m)^\top &
        (\delta u_{k-1}^m)^\top
    \end{bmatrix}^{\top}
\end{equation}
Stacking the regressors and corresponding next-sample differential outputs as columns of $\mathcal X$ and $\mathcal Y$, respectively, the model matrices are estimated jointly using ridge regression
\begin{equation*}
    \Theta^\star=\operatorname*{arg\,min}_{\Theta}\left\|\mathcal Y-\Theta\mathcal X\right\|_F^2+\alpha\left\|\Theta\right\|_F^2, \quad
    \Theta=\begin{bmatrix}A&B&C\end{bmatrix}
\end{equation*}
where $\alpha\geq0$ is the regularization parameter. Notice that the affine offset $d$ does not need to be estimated since it vanishes in the differential representation.

Given a candidate command correction $\Delta U_j$, the identified model predicts the corresponding change in measured motion according to
\begin{align*}
    \Delta y_0 &= 0 \label{eq:incremental-initial}\\
    \Delta y_{k+1} &= A\Delta y_k+B\Delta u_k+C\Delta u_{k-1}, \quad k=0,\ldots,N-2
\end{align*}
with $\Delta u_{-1}=0$. The predicted next-trial trajectory is
\begin{equation*}
    \widehat Y_{j+1}=Y_j+\Delta Y_j
\end{equation*}
Thus, the identified model predicts only the change induced by the candidate command correction, while the trajectory prediction is re-anchored at the measured robot motion after every physical trial.

The differential model is identified once prior to the ILC experiments and subsequently kept fixed. No model re-identification is performed during the trial-to-trial learning process. As shown later in the experiments, this single compact model is sufficient to support effective constrained model-based ILC across the considered repetitive motion tasks and operating conditions.
\vspace{-2mm}

\begin{figure*}[!h]
    \centering 
    \includegraphics[width=\textwidth, height=6cm]{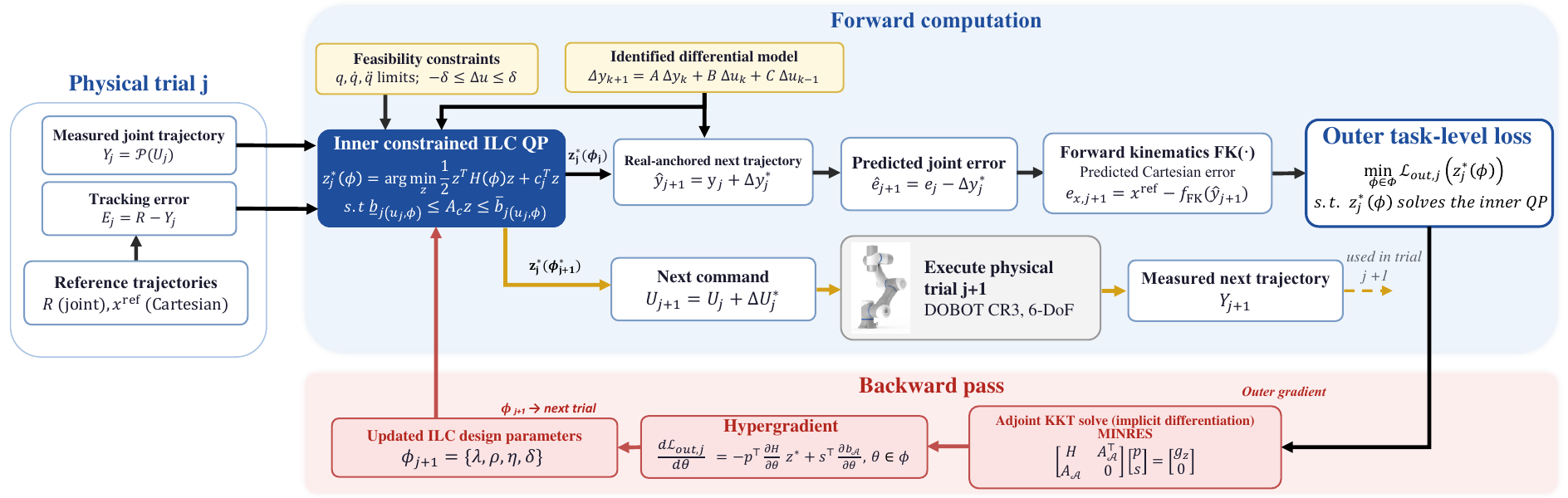} \caption{Computational graph of \DiffILC{} at physical trial $j$. The measured trajectory anchors the constrained ILC update and real-anchored next-trial prediction, while hypergradients through the locally active KKT system update the ILC parameters over multiple computational epochs. After the final outer update, the inner QP is re-solved using $\phi_{j+1}$, and only the resulting updated command is executed on the robot at trial $j+1$.}\label{fig:diff-ilc-framework}
    \vspace{-5mm}
\end{figure*}

\vspace{-2mm}
\section{Differentiable Constrained Iterative Learning Control} \label{sec:Diff-ILC}

\subsection{Constrained Model-Based ILC} \label{subsec:constrained-ilc}
At trial $j$, the measured trajectory $Y_j$ and tracking error $E_j$ are
available from the physical execution of the robot. Using the differential
model developed in \autoref{subsec:differential-model}, we seek a command
correction $\Delta U_j$ whose predicted motion change reduces the measured
tracking error while maintaining a smooth and feasible command trajectory.

For the optimization formulation, let $y_j$, $e_j$, $u_j$, $\Delta y$, and
$\Delta u$ denote the time-major stacked forms of $Y_j$, $E_j$, $U_j$,
$\Delta Y$, and $\Delta U$, respectively, with subscript $k$ denoting the
corresponding time-step block. Let $D_1$ and $D_2$ denote the first- and
second-order finite-difference operators over the trajectory horizon. Define
the normalization constants
$n_e=(N-1)n_q$, $n_u=Nn_q$, $n_1=(N-1)n_q$, and $n_2=(N-2)n_q$.
Then, the constrained model-based ILC update is formulated as
\begin{align}
(\Delta y_j^\star,\Delta u_j^\star)
&=\operatorname*{arg\,min}_{\Delta y,\Delta u}
\frac{1}{2n_e}\|e_j-\Delta y\|_{\bar Q}^{2}
+\frac{1}{2n_u}\|\Delta u\|_{\bar\Lambda}^{2}
\nonumber\\
&\quad
+\frac{1}{2n_1}\|D_1\Delta u\|_{\bar P_\rho}^{2}
+\frac{1}{2n_2}\|D_2\Delta u\|_{\bar P_\eta}^{2}
\label{eq:inner-problem}\\
\text{s.t.}\qquad
\Delta y_0 &= 0
\nonumber\\
 \Delta y_{k+1}
=&A\Delta y_k+B\Delta u_k+C\Delta u_{k-1},
\quad k=0,\cdots,N-2
\nonumber\\
q_{\min}
&\leq u_j+\Delta u\leq q_{\max}
\nonumber\\
-\delta
&\leq \Delta u\leq\delta
\nonumber\\
-hv_{\max}
&\leq D_1(u_j+\Delta u)\leq hv_{\max}
\nonumber\\
-h^2a_{\max}
&\leq D_2(u_j+\Delta u)\leq h^2a_{\max}
\nonumber
\end{align}
where the joint-wise limits are understood to be repeated over the trajectory horizon and $\Delta u_{-1}=0$. The quadratic terms are normalized according to the trajectory horizon and the corresponding finite-difference order, reducing the dependence of their relative weighting on the horizon length.

The first term in the objective in \eqref{eq:inner-problem} penalizes the
predicted residual tracking error $e_j-\Delta y$. The remaining terms
regularize the magnitude, temporal variation, and curvature of the command
correction, respectively. Together with the physical constraints in
\eqref{eq:inner-problem}, these terms determine the aggressiveness and
smoothness of the learning update.

The weighting matrices are constructed from joint-specific regularization parameters. Let $ \Lambda=\operatorname{diag}(\lambda_1,\ldots,\lambda_{n_q}), \ P_\rho=\operatorname{diag}(\rho_1,\ldots,\rho_{n_q})$ and $P_\eta=\operatorname{diag}(\eta_1,\ldots,\eta_{n_q})$ and define their horizon extensions as $\bar\Lambda=I_N\otimes\Lambda, \  \bar P_\rho=I_{N-1}\otimes P_\rho,   \  \bar P_\eta=I_{N-2}\otimes P_\eta$, where $\otimes$ denotes the Kronecker product. Similarly, for a fixed joint-space tracking metric $Q\succeq0$, we define $\bar Q=I_N\otimes Q$.

We keep the tracking metric $Q$ fixed and adapt the regularization and correction parameters. This removes the common scale ambiguity that would arise from jointly scaling all quadratic weights while retaining the relative trade-off between tracking improvement, correction magnitude, and temporal smoothness. The parameters governing the ILC update are therefore collected as
\begin{equation}
    \phi=
    \begin{bmatrix}
        \lambda_1,\ldots,\lambda_{n_q},
        \rho_1,\ldots,\rho_{n_q},
        \eta_1,\ldots,\eta_{n_q},
        \delta
    \end{bmatrix}^{\top}
    \label{eq:ilc-parameter-vector}
\end{equation}

For a fixed $\phi$, the optimization problem in \eqref{eq:inner-problem} defines a conventional constrained model-based ILC update, i.e. the QP realizes the parameterized learning law in \eqref{eq:update-policy}, with
$\pi(E_j,U_j;\phi)=\Delta U_j^\star(\phi)$. The command applied on the following physical trial is $U_{j+1}=U_j+\Delta U_j^\star(\phi)$. 

For the subsequent sensitivity analysis, we write \eqref{eq:inner-problem} in standard quadratic-program form. Define $z = [ \Delta y^\top \ \  \Delta u^\top]^{\top}$. Then the constrained ILC update can be expressed as
\begin{align} \label{eq:standard-qp}
z_j^\star(\phi) & =\operatorname*{arg\,min}_{z}
\frac{1}{2}z^\top H(\phi)z+c_j^\top z\\ 
&\text{s.t.}\quad
\underline b_j(u_j,\phi) \leq A_cz\leq \overline b_j(u_j,\phi)
\end{align}
where $c_j = [ -\dfrac{1}{n_e}\bar Q e_j \ \ \  0 ]^\top$ and
\vspace{-1mm}
\begin{equation*}
H(\phi) = \operatorname{blkdiag} ( \frac{1}{n_e}\bar Q,\,
\frac{1}{n_u}\bar\Lambda + \frac{1}{n_1}D_1^\top\bar  P_\rho D_1 + \frac{1}{n_2}D_2^\top\bar P_\eta D_2  )
\end{equation*}
where $\operatorname{blkdiag}(\cdot)$ denotes block-diagonal assembly. The matrix $A_c$ encodes the differential dynamics and command constraints, with corresponding bounds $\underline b_j(u_j,\phi)$ and $\overline b_j(u_j,\phi)$. The parameters $\lambda_i$, $\rho_i$, and $\eta_i$ enter $H(\phi)$, whereas $\delta$ enters the bounds. Retaining both $\Delta y$ and $\Delta u$ as decision variables preserves the
sparse finite-horizon structure induced by the differential dynamics, whereas eliminating $\Delta y$ would result in a condensed QP in $\Delta u$ with generally dense quadratic terms. Following~\cite{meindl2025iterative}, we refer to this data-driven formulation as model-based ILC. The next subsection treats $\phi$ as learnable and adapts it between physical trials.

% with $\operatorname{blkdiag}(\cdot)$ denoting the block-diagonal matrix formed from its arguments. The matrix $A_c$ collects the differential dynamics and command constraints in \eqref{eq:inner-problem}, while $\underline b_j(u_j,\phi)$ and $\overline b_j(u_j,\phi)$ collect the corresponding lower and upper bounds. The regularization parameters $\lambda_i$, $\rho_i$, and $\eta_i$ enter the Hessian, whereas the correction limit $\delta$ enters the constraint bounds.

% We retain both $\Delta y$ and $\Delta u$ as optimization variables rather than eliminating the differential states. This preserves the block-banded structure induced by the next-trial dynamics and finite-difference operators, avoiding the dense quadratic terms that arise in a condensed lifted formulation.
%  Following~\cite {meindl2025iterative}, we refer to the data-driven ILC as model-based ILC and use the term model-based ILC throughout this paper.

% The following subsection treats $\phi$ as a set of learnable parameters and develops an outer-level objective for adapting them between physical trials.
\vspace{-3mm}
\subsection{Bilevel Parameter Adaptation}
\label{subsec:bilevel-adaptation}
The correction quality depends on the regularization parameters and correction limit, which balance predicted tracking improvement against correction magnitude, temporal variation, and curvature. We therefore adapt these parameters at the outer level based on the solution of the inner ILC problem. \autoref{fig:diff-ilc-framework} summarizes the resulting trial-level computational graph, including the real-anchored prediction, outer parameter update, and subsequent physical execution.

% The constrained quadratic program in \eqref{eq:standard-qp} produces the command correction that will be applied on the next physical trial. However, the quality of this correction depends on the choice of the regularization parameters and the correction limit. These parameters govern the trade-off between predicted tracking improvement, command magnitude, temporal variation, and curvature. We therefore adapt them at an outer level using the outcome of the inner ILC problem.

For a fixed parameter vector $\phi$, the inner solution predicts the joint trajectory on the next trial as
\begin{equation} \label{eq:predicted-output}
    \widehat y_{j+1}(\phi) = y_j+\Delta y_j^\star(\phi)
\end{equation}
Consequently, the predicted joint-space tracking error is
\begin{equation} \label{eq:predicted-joint-error}
    \widehat e_{j+1}(\phi) = e_j-\Delta y_j^\star(\phi)
\end{equation}
It is worth noting that \eqref{eq:predicted-output} is anchored at the measured trajectory $y_j$ obtained from the current physical execution. The identified model is used only to predict the incremental effect of the candidate ILC correction rather than to generate an absolute rollout of the robot motion. A new physical measurement re-anchors this prediction after every ILC trial.

Although the inner problem is formulated in joint space, the tracking performance can also be evaluated directly in Cartesian task space. Let $f_{\mathrm{FK}}:\mathbb{R}^{n_q}\rightarrow\mathbb{R}^{n_x}$ denote a differentiable forward-kinematics map, and let $x_k^{\mathrm{ref}}\in\mathbb{R}^{n_x}$ denote the desired Cartesian reference at time step $k$. The Cartesian trajectory predicted by the differential model is obtained by applying forward kinematics to the predicted joint trajectory. Hence, the predicted Cartesian tracking error is
\begin{equation*}
    e_{x,j+1,k}(\phi) = x_k^{\mathrm{ref}} - f_{\mathrm{FK}}(y_{j,k}+\Delta y_{j,k}^\star(\phi)), \quad
    k=0,\ldots,N-1
\end{equation*}
Stacking these errors over the trajectory horizon gives
\begin{equation}
    e_{x,j+1}(\phi) = \operatorname{col}(e_{x,j+1,0}^\top,\ldots,e_{x,j+1,N-1}^\top)^\top
    \label{eq:stacked-cartesian-error}
\end{equation}
Since $a_{\max}\in\mathbb{R}^{n_q}$ contains the joint-wise acceleration limits, we define the horizon-level normalization matrix $\Gamma_a = I_{N-2}\otimes \operatorname{diag}(\frac{1}{h^2a_{\max}})$, where the reciprocal is applied element-wise. For the Cartesian tracking term, define the normalization constant $n_c=Nn_q$. The outer objective evaluates whether the correction generated by the inner problem is desirable from the perspective of the next physical trial. We define
\begin{equation}
\begin{aligned}
    \mathcal L_{\mathrm{out},j}(z_j^\star(\phi))
    = & \frac{w_q}{2n_e}\left\|\widehat e_{j+1}(\phi)\right\|_2^2
    +\frac{w_x}{2n_c}\left\|e_{x,j+1}(\phi)\right\|_2^2\\
    &+\frac{w_a}{2n_2}\left\|\Gamma_aD_2(u_j+\Delta u_j^\star(\phi))\right\|_2^2
\end{aligned}
\label{eq:outer-loss}
\end{equation}
where $w_q$, $w_x$, and $w_a$ are fixed outer-level weights. The first term evaluates the predicted next-trial tracking error in joint space, while the second evaluates the same predicted correction in Cartesian task space. The final term discourages unnecessary approach to the acceleration limits and promotes a smooth updated command trajectory. The outer weights define the task-level evaluation criterion and are kept fixed; they are not included in the learnable parameter vector $\phi$.

The resulting parameter-learning problem is bilevel because the outer objective depends on the solution of the constrained inner problem
\begin{equation}
\begin{aligned}
    \operatorname*{minimize}_{\phi\in\Phi}\quad
    &\mathcal L_{\mathrm{out},j}\left(z_j^\star(\phi)\right)\\
    \text{subject to}\quad
    &z_j^\star(\phi)\text{ solves the inner QP in }\eqref{eq:standard-qp}
\end{aligned}
\label{eq:bilevel-problem}
\end{equation}
where $\Phi$ denotes the admissible set of ILC parameters defined in Section~\ref{subsec:parameterization-learning}. During the outer optimization, the measured trial data $e_j$ and $u_j$ are held fixed. Only the parameters governing the correction to be applied on the next physical trial are adapted.
\vspace{-3mm}
\subsection{Implicit Differentiation Through Constrained ILC Update}
\label{subsec:implicit-differentiation}

The outer objective depends on $\phi$ through the optimizer $z_j^\star(\phi)$. Rather than differentiating through the iterations of a particular numerical QP solver, we differentiate the local optimality conditions that characterize the constrained ILC solution.

To first compute the gradient of the outer objective with respect to the primal QP variables, define the block-diagonal forward-kinematics Jacobian
\begin{equation}
    J_{\mathrm{FK}}
    =
    \operatorname{blkdiag}
    \left(
        J_{\mathrm{FK},0},
        \ldots,
        J_{\mathrm{FK},N-1}
    \right)
    \label{eq:fk-jacobian}
\end{equation}
where $J_{\mathrm{FK},k} = \left. \frac{\partial f_{\mathrm{FK}}(q)} {\partial q} \right|_{q=y_{j,k}+\Delta y_{j,k}^\star}$. The gradient of the outer loss with respect to the inner solution is then
\begin{equation}
    g_z
    \triangleq
    \left.
    \nabla_z
    \mathcal L_{\mathrm{out},j}
    \right|_{z=z_j^\star}
    =
    \begin{bmatrix}
        g_y\\
        g_u
    \end{bmatrix}
    \label{eq:outer-solution-gradient}
\end{equation}
with $g_y = -\frac{w_q}{n_e}\widehat e_{j+1} - \frac{w_x}{n_c} J_{\mathrm{FK}}^\top e_{x,j+1}$ and $g_u = \frac{w_a}{n_2} D_2^\top \Gamma_a^\top\Gamma_a D_2 ( u_j+\Delta u_j^\star)$. These expressions show explicitly how the joint-space, Cartesian-space, and command-smoothness criteria propagate to the variables of the constrained ILC problem.

Let $\mathcal A$ contain all equality constraints and all inequality constraints that are active at $z_j^\star$. The locally active constraints can be written as $A_{\mathcal A}z_j^\star = b_{\mathcal A}(u_j,\phi)$, where $b_{\mathcal A}$ contains the corresponding equality values and active lower or upper bounds. The Lagrangian associated with the locally active QP is
\begin{equation}
    \mathcal L_{\mathrm{QP}}
    =
    \frac{1}{2}z^\top H(\phi)z
    +
    c_j^\top z
    +
    \nu_{\mathcal A}^\top
    \left(
        A_{\mathcal A}z
        -
        b_{\mathcal A}(u_j,\phi)
    \right)
    \label{eq:local-qp-lagrangian}
\end{equation}
The corresponding KKT conditions are
\begin{align}
    H(\phi)z_j^\star
    +c_j
    +A_{\mathcal A}^\top\nu_{\mathcal A}^\star
    &=0
    \label{eq:kkt-stationarity}\\
    A_{\mathcal A}z_j^\star
    -b_{\mathcal A}(u_j,\phi)
    &=0
    \label{eq:kkt-feasibility}
\end{align}

Consider a scalar parameter $\theta$ contained in $\phi$. During the outer update, the measured trial data are fixed, and therefore $c_j$ and $A_{\mathcal A}$ do not depend on $\theta$. In a neighborhood where the active set remains unchanged, differentiating \eqref{eq:kkt-stationarity}--\eqref{eq:kkt-feasibility} gives
\begin{equation}
    \begin{bmatrix}
        H & A_{\mathcal A}^\top\\
        A_{\mathcal A} & 0
    \end{bmatrix}
    \begin{bmatrix}
        \dfrac{\partial z_j^\star}{\partial\theta}\\[1mm]
        \dfrac{\partial\nu_{\mathcal A}^\star}{\partial\theta}
    \end{bmatrix}
    =
    \begin{bmatrix}
        -\dfrac{\partial H}{\partial\theta}z_j^\star\\[1mm]
        \dfrac{\partial b_{\mathcal A}}{\partial\theta}
    \end{bmatrix}
    \label{eq:forward-kkt-sensitivity}
\end{equation}
The sensitivity analysis is local. We assume that the active set remains unchanged in a neighborhood of the current solution and that the active constraints satisfy the linear-independence constraint qualification, with a nonsingular local KKT system. Under these standard regularity conditions, the constrained optimizer defines a locally differentiable solution map.

Computing \eqref{eq:forward-kkt-sensitivity} separately for every component of $\phi$ would require one KKT solve per parameter. We instead introduce the primal and active-constraint adjoints $p$ and $s$ through
\begin{equation}
    \begin{bmatrix}
        H & A_{\mathcal A}^\top\\
        A_{\mathcal A} & 0
    \end{bmatrix}
    \begin{bmatrix}
        p\\
        s
    \end{bmatrix}
    =
    \begin{bmatrix}
        g_z\\
        0
    \end{bmatrix}
    \label{eq:adjoint-kkt}
\end{equation}
The adjoint system involves the same symmetric KKT matrix as the forward sensitivity problem. The hypergradient of the outer objective with respect to any scalar parameter $\theta\in\phi$ is therefore
\begin{equation}
    \frac{d\mathcal L_{\mathrm{out},j}}{d\theta}
    =
    -p^\top
    \frac{\partial H}{\partial\theta}
    z_j^\star
    +
    s^\top
    \frac{\partial b_{\mathcal A}}{\partial\theta}
    \label{eq:adjoint-hypergradient}
\end{equation}
Thus, a single adjoint KKT solve provides the quantities required to evaluate the hypergradients of all learnable ILC parameters.

To make this dependence explicit, partition $p=[p_y^\top\;p_u^\top]^\top$ and define $S_i=\mathbf e_i\mathbf e_i^\top$, where $\mathbf e_i$ is the $i$th standard basis vector in $\mathbb{R}^{n_q}$. The joint-specific regularization gradients are
\begin{align}
    \frac{d\mathcal L_{\mathrm{out},j}}{d\lambda_i}
    &=
    -\frac{1}{n_u}
    p_u^\top
    \left(
        I_N\otimes S_i
    \right)
    \Delta u_j^\star
    \label{eq:lambda-gradient}\\
    \frac{d\mathcal L_{\mathrm{out},j}}{d\rho_i}
    &=
    -\frac{1}{n_1}
    \left(
        D_1p_u
    \right)^\top
    \left(
        I_{N-1}\otimes S_i
    \right)
    D_1\Delta u_j^\star
    \label{eq:rho-gradient}\\
    \frac{d\mathcal L_{\mathrm{out},j}}{d\eta_i}
    &=
    -\frac{1}{n_2}
    \left(
        D_2p_u
    \right)^\top
    \left(
        I_{N-2}\otimes S_i
    \right)
    D_2\Delta u_j^\star
    \label{eq:eta-gradient}
\end{align}

The correction limit $\delta$ enters through the constraint bounds rather than the QP Hessian. Let $\mathcal A_\delta^+$ and $\mathcal A_\delta^-$ denote the active upper and lower correction-bound rows, respectively. Its hypergradient is
\begin{equation}
    \frac{d\mathcal L_{\mathrm{out},j}}{d\delta}
    =
    \sum_{r\in\mathcal A_\delta^+}s_r
    -
    \sum_{r\in\mathcal A_\delta^-}s_r
    \label{eq:delta-gradient}
\end{equation}
When no correction-bound row is active, the local derivative with respect to $\delta$ is zero.
\vspace{-4mm}
\subsection{Constrained Parameterization and Trial-Level Learning}
\label{subsec:parameterization-learning}

The learnable parameters must remain within their admissible ranges during the outer optimization. We therefore optimize unconstrained raw parameters and map them to valid ILC parameters through differentiable transformations.

For each positive regularization parameter
$w_i\in\{\lambda_i,\rho_i,\eta_i\}$, let $a_{w_i}\in\mathbb{R}$ denote its corresponding raw parameter and define $w_i= \underline w_i + \operatorname{softplus} (a_{w_i})$, where $\underline w_i\geq0$ is a prescribed lower bound. The shared correction limit is constrained to a prescribed interval by $\delta = \delta_{\min} +
    ( \delta_{\max}-\delta_{\min}) \sigma(a_\delta)$.
Here, $\sigma(\cdot)$ denotes the logistic sigmoid, and the transformations collectively define $\phi=\mathcal T(a)$. The hypergradients are propagated through $\mathcal T$ by the chain rule, and the raw parameters are updated using Adam. At each nonterminal physical trial, the measured trajectory is held fixed while $K_{\mathrm{out}}$ outer epochs update the parameters computationally. The inner QP is then re-solved with the final parameters, and only the resulting correction $\Delta U_j^\star$ is applied on the next physical trial, where $\Delta U_j^\star$ is the trajectory form of $\Delta u_j^\star$. Algorithm~\ref{alg:differentiable-ilc} summarizes the complete procedure.

% where $\sigma(\cdot)$ denotes the logistic sigmoid. Collectively, these transformations define $\phi=\mathcal T(a)$. The hypergradients obtained in Section~\ref{subsec:implicit-differentiation} are propagated through $\mathcal T$ by the chain rule, and the raw parameters are updated using Adam.

% At each nonterminal physical trial, the measured trajectory and tracking error define the current inner ILC problem. These measured quantities are held fixed while several outer optimization epochs are performed. After the final parameter update, the inner QP is solved once more using the updated parameters, and only this final correction is applied to the robot on the subsequent physical trial. Algorithm~\ref{alg:differentiable-ilc} summarizes the complete procedure. Note that $\Delta U_j^\star$ denotes the trajectory form of the vectorized optimizer $\Delta u_j^\star$.

\begin{algorithm}[h]
\caption{Differentiable Iterative Learning Control}
\label{alg:differentiable-ilc}
\footnotesize
\begin{algorithmic}[1]

\Require Reference trajectory $R$; identified differential model $(A,B,C)$; feasible initial command $U_1$; initial raw parameters $a_1$; number of physical trials $J$; number of outer epochs $K_{\mathrm{out}}$
\Ensure Learned command $U_J$ and parameters $\phi_J$
\For{$j=1,\ldots,J$}
    \State Execute $U_j$ on the robot, measure $Y_j$, and compute
    $E_j\gets R-Y_j$
    \If{$j<J$}
        \State $a\gets a_j$
        \For{$m=1,\ldots,K_{\mathrm{out}}$}
            \State $\phi\gets\mathcal T(a)$
            \State Solve the inner QP \eqref{eq:standard-qp} for
            $z_j^\star(\phi)$
            \State Evaluate the outer loss \eqref{eq:outer-loss}
            and compute $g_z$
            \State Determine the local active set $\mathcal A$
            \State Solve the adjoint KKT system
            \eqref{eq:adjoint-kkt}
            \State Evaluate
            $\nabla_\phi\mathcal L_{\mathrm{out},j}$
            using
            \eqref{eq:adjoint-hypergradient}--\eqref{eq:delta-gradient}
            \State Propagate the hypergradient through
            $\phi=\mathcal T(a)$
            \State Update $a$ using Adam
        \EndFor
        \State $\phi_{j+1}\gets\mathcal T(a)$
        \State Re-solve \eqref{eq:standard-qp} using $\phi_{j+1}$
        \State $U_{j+1}\gets U_j+\Delta U_j^\star$
        \State $a_{j+1}\gets a$
    \EndIf
\EndFor
\State $\phi_J\gets\mathcal T(a_J)$
\State \Return $U_J,\phi_J$
\end{algorithmic}
\end{algorithm}
% \vspace{-5mm}
% The resulting procedure separates learning across two time scales. Each physical execution provides a new measured trajectory, after which $K_{\mathrm{out}}$ parameter updates are performed computationally without additional robot executions. Only the correction obtained after the final outer update is applied on the subsequent physical trial. This separation allows the parameters governing the ILC update to adapt while directly limiting the number of physical repetitions required for learning.
\vspace{-2mm}
\section{Experimentation and Results}
\label{sec:experiments}

\subsection{Experimental Setup}

Experiments were conducted on a 6-DoF DOBOT CR3 using only its joint-position command interface. The ILC experiments operated at $100~\mathrm{Hz}$. For model identification, data were acquired at $125~\mathrm{Hz}$ and resampled to the control grid. Three nominal executions and six symmetric smooth multi-joint perturbation directions were collected about a $0.05~\mathrm{Hz}$ joint-space sinusoidal trajectory, resulting in $15$ physical trials. The perturbations were generated using $10$ B-spline basis functions and bounded by $0.5^\circ$ in position, $2^\circ/\mathrm{s}$ in velocity, and $10^\circ/\mathrm{s}^2$ in acceleration. The matrices $(A,B,C)$ were estimated using ridge regression with $\alpha=0.01$ and then kept fixed throughout all ILC experiments; no re-identification was performed for different trajectories or execution speeds.

Each ILC experiment comprised $10$ physical trials. The fixed-parameter constrained model-based ILC baseline used $\lambda_i=\rho_i=\eta_i=1$ for all six joints and $\delta=10^\circ$. \DiffILC{} was initialized with the same regularization parameters, but with $\delta=1^\circ$, which was adapted within $0.05^\circ\leq\delta\leq10^\circ$. The regularization parameters remained positive through the softplus parameterization. Both methods used the same differential model, tracking metric $Q=I_6$, initial command, and joint-position, velocity, and acceleration constraints. Joint positions were limited to $\pm350^\circ$ for Joints~1, 2, 4, 5, and 6 and to $\pm150^\circ$ for Joint~3; command velocity and acceleration were limited to $100^\circ/\mathrm{s}$ and $300^\circ/\mathrm{s}^2$, respectively. Two \DiffILC{} configurations were evaluated using five and ten outer optimization epochs between physical executions.

The inner QP was solved using OSQP \cite{stellato2020osqp}, and the adjoint KKT system was solved using MINRES \cite{paige1975minres}. The losses, forward kinematics, and parameter transformations were implemented in double-precision PyTorch, and the raw parameters were updated using Adam with learning rate $0.2$. The outer-loss weights were fixed at $w_q=1$, $w_x=0.2$, and $w_a=2$. The adjoint hypergradient was verified against finite differences for all $19$ learnable parameters at multiple physical trials. Close agreement was obtained whenever the local active set remained unchanged.

The evaluation used unseen planar circle, infinity, and star trajectories and a three-dimensional helix. The circle and helix had radius $150~\mathrm{mm}$, with a helix pitch of $50~\mathrm{mm}$. The infinity trajectory had width $400~\mathrm{mm}$ and height $150~\mathrm{mm}$, while the star had outer radius $150~\mathrm{mm}$ and inner-to-outer radius ratio $0.382$. Cartesian paths were converted to joint-space references using inverse kinematics with fixed end-effector orientation and then smoothed. The four primary trajectories were executed at $0.1~\mathrm{Hz}$ for $N=2001$ samples, corresponding to approximately $20~\mathrm{s}$. An additional circle was executed at $0.15~\mathrm{Hz}$ to evaluate transfer to faster motion. All applied commands satisfied the prescribed constraints within the numerical feasibility tolerance of the QP solver. Tracking performance is evaluated using the RMS over time of the joint-error norm,
$\sqrt{\frac{1}{N}\sum_{k=0}^{N-1}\|e_{j,k}\|_2^2}$, and analogously the Cartesian position-error norm.

A supplementary video demonstrating the robot experiments
and tracking results is available at
\url{https://www.youtube.com/watch?v=Jrib0FHbBd4}.

\vspace{-5mm}
\subsection{Tracking Convergence}
% \vspace{-3mm}
\begin{figure}[t]
    \centering
    \includegraphics[width=\linewidth]{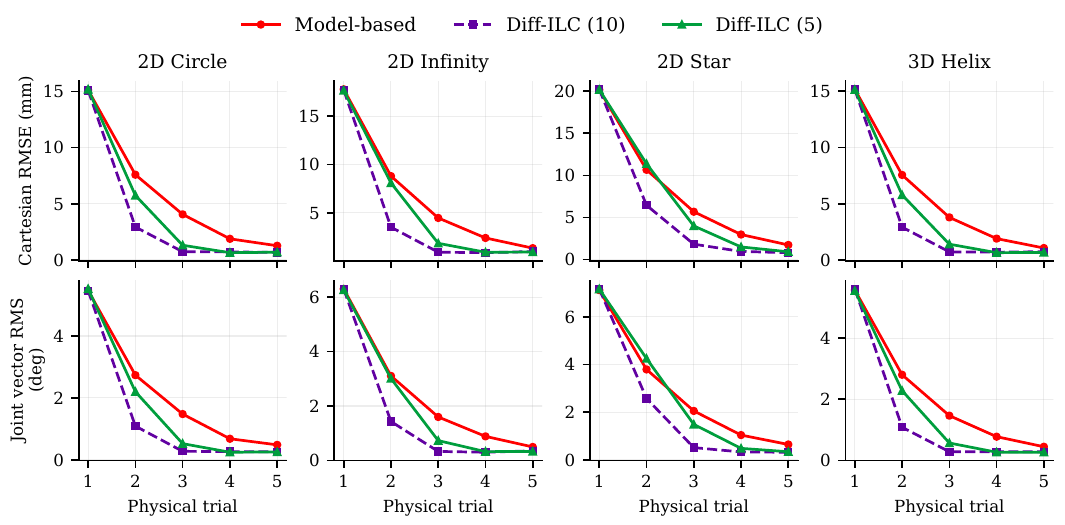}
    \caption{Joint and Cartesian tracking convergence over the first five physical trials for the four $0.1~\mathrm{Hz}$ evaluation trajectories using fixed-parameter model-based ILC and \DiffILC{} with five and ten outer optimization epochs between physical executions.}
    \label{fig:convergence}
    \vspace{-4mm}
\end{figure}

As shown in \autoref{fig:convergence}, all methods improve tracking across the four unseen trajectories. The average joint/Cartesian RMSE decreases from $6.13^\circ/17.05~\mathrm{mm}$ at Trial~1 to $1.65^\circ/4.50~\mathrm{mm}$ for fixed-parameter model-based ILC by Trial~3. 

Over the same trials, \DiffILC{} reaches $0.83^\circ/2.14~\mathrm{mm}$ with five epochs and $0.35^\circ/1.06~\mathrm{mm}$ with ten epochs; the latter corresponds to $78.8\%$ and $76.4\%$ lower joint and Cartesian RMSE, respectively, than the fixed baseline. By Trial~5, both \DiffILC{} configurations achieve approximately $0.30^\circ/0.81~\mathrm{mm}$. Notably, by Trial~3, after only two trial-to-trial updates \DiffILC{} already achieves lower mean error than the fixed-parameter baseline does by Trial~5, directly demonstrating the gain in physical-trial efficiency.

\begin{table}[h]
\vspace{-4mm}
\centering
\setlength{\abovecaptionskip}{2pt}
\setlength{\belowcaptionskip}{2pt}
\caption{\scriptsize Joint-vector and Cartesian tracking performance at selected physical trials. Entries report the mean joint-vector RMS ($^\circ$)/Cartesian position RMS (mm) over three repeated experiments. The penultimate column reports the trial pair $T_{\mathrm{MB}}/T_{\mathrm{Diff}}$ at which fixed MB-ILC reaches the 5\% criterion relative to Diff-ILC for both joint and Cartesian RMS errors; the final column gives the Trial~3 reduction relative to fixed-parameter model-based ILC.}
\label{tab:combined-convergence}
\setlength{\tabcolsep}{4.5pt}
\renewcommand{\arraystretch}{1.10}

\resizebox{\columnwidth}{!}{%
\begin{tabular}{llccccc}
\toprule
Trajectory & Method & T1 (nominal) & T3 & T5
& $T_{\mathrm{MB}}/T_{\mathrm{Diff}}$ (5\%)
& $\Delta_{\mathrm{T3}}$ J/C (\%) \\
\midrule

\multirow{3}{*}{Circle}
& Fixed MB-ILC
& 5.47/15.12 & 1.48/4.05 & 0.49/1.26 & -- & -- \\
& \DiffILC{} (5 epochs)
& 5.52/15.18 & 0.53/1.30 & 0.26/0.68 & T5/T3 & 64.2/67.9 \\
& \DiffILC{} (10 epochs)
& 5.46/15.09 & \textbf{0.28/0.73} & \textbf{0.26/0.67}
& T7/T3 & \textbf{81.1/82.0} \\
\midrule

\multirow{3}{*}{Fast circle}
& Fixed MB-ILC
& 9.06/22.60 & 2.40/6.02 & 0.84/2.06 & -- & -- \\
& \DiffILC{} (5 epochs)
& 9.11/22.73 & 2.10/4.62 & 0.55/1.30 & T4/T3 & 12.5/23.3 \\
& \DiffILC{} (10 epochs)
& 9.06/22.59 & \textbf{0.60/1.45} & 0.53/1.26
& T8/T3 & \textbf{75.0/75.9} \\
\midrule

\multirow{3}{*}{Helix}
& Fixed MB-ILC
& 5.57/15.16 & 1.46/3.79 & 0.45/1.07 & -- & -- \\
& \DiffILC{} (5 epochs)
& 5.55/15.12 & 0.57/1.41 & 0.27/0.68 & T5/T3 & 61.0/62.8 \\
& \DiffILC{} (10 epochs)
& 5.61/15.15 & \textbf{0.28/0.73} & 0.28/0.74
& T7/T3 & \textbf{80.8/80.7} \\
\midrule

\multirow{3}{*}{Infinity}
& Fixed MB-ILC
& 6.31/17.77 & 1.60/4.48 & 0.49/1.36 & -- & -- \\
& \DiffILC{} (5 epochs)
& 6.27/17.65 & 0.73/1.86 & 0.33/0.96 & T5/T3 & 54.4/58.5 \\
& \DiffILC{} (10 epochs)
& 6.31/17.71 & \textbf{0.33/0.95} & 0.35/1.01
& T8/T3 & \textbf{79.4/78.8} \\
\midrule

\multirow{3}{*}{Star}
& Fixed MB-ILC
& 7.12/20.18 & 2.06/5.68 & 0.65/1.72 & -- & -- \\
& \DiffILC{} (5 epochs)
& 7.17/20.19 & 1.49/3.99 & 0.35/0.91 & T4/T3 & 27.7/29.8 \\
& \DiffILC{} (10 epochs)
& 7.17/20.23 & \textbf{0.52/1.82} & 0.33/0.81
& T6/T3 & \textbf{74.8/68.0} \\

\bottomrule
\end{tabular}%
}
\end{table}

Increasing the number of outer epochs can improve early physical-trial convergence, but at higher computational cost. The ten-epoch configuration performs best at Trial~3, whereas by Trial~5 the five- and ten-epoch configurations achieve similar performance, with five epochs marginally better for the helix and infinity trajectories. Because all outer epochs optimize the same real-anchored next-trial surrogate constructed from the fixed identified differential model, additional optimization can improve the predicted update without necessarily yielding proportional improvement on the subsequent physical trial under model mismatch. Re-anchoring the prediction at the measured trajectory after every physical execution prevents such mismatch from accumulating across trials. 
% Increasing the number of outer optimization epochs primarily improves early convergence in terms of physical repetitions. The ten-epoch configuration produces the lowest average tracking error by Trial~3, whereas by Trial~5 the five- and ten-epoch configurations achieve nearly identical average performance. The ten-epoch configuration is also not uniformly superior at Trial~5; for example, the five-epoch configuration yields marginally lower errors for the helix and infinity trajectories. Increasing the number of outer epochs therefore yields diminishing physical returns while increasing the computation performed between robot executions. The average parameter-tuning times were $48.8~\mathrm{s}$ and $113.1~\mathrm{s}$ for the five- and ten-epoch configurations, respectively.

% This behavior is consistent with the structure of the bilevel update. The outer epochs optimize a real-anchored next-trial surrogate constructed from the fixed identified differential model rather than the subsequent physical tracking loss itself. Additional epochs can therefore improve the parameter update before the next robot execution, but they also increase the extent to which the learned parameters are optimized against the surrogate. Consequently, more outer optimization does not necessarily translate into proportional improvement on the subsequent physical execution when model mismatch is present.

\begin{figure}[h]
\vspace{-3mm}
    \centering
    \includegraphics[width=\linewidth]{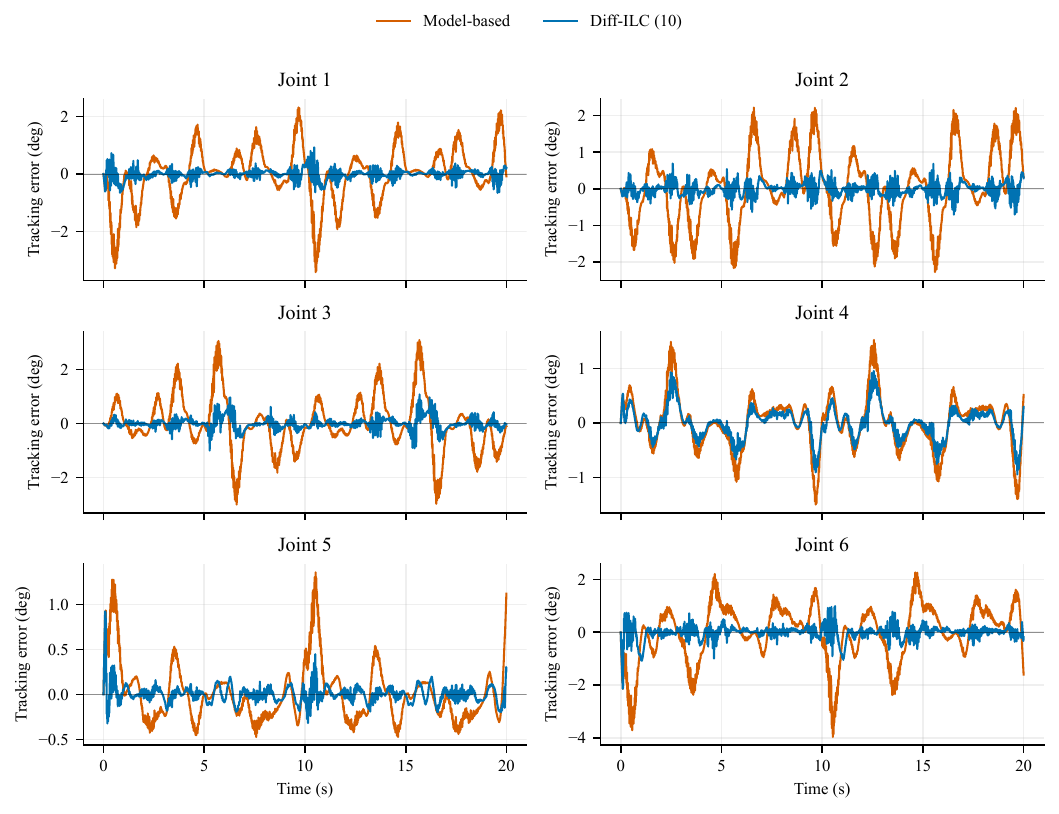}
    \caption{Joint-wise tracking errors for the star trajectory at Trial~3 using fixed-parameter model-based ILC and \DiffILC{} with ten outer optimization epochs.}
    \label{fig:star-trial3-joint-errors}
    \vspace{-4mm}
\end{figure}

For the challenging star trajectory, \autoref{fig:star-trial3-joint-errors} shows that \DiffILC{} reduces the Trial~3 joint RMSE from $2.06^\circ$ to $0.52^\circ$ and the Cartesian RMSE from $5.68$ to $1.82~\mathrm{mm}$, corresponding to reductions of $74.8\%$ and $68.0\%$, respectively. Although the star’s sharp direction changes produce larger residual errors, \DiffILC{} achieves substantial improvement within three physical trials without trajectory-specific retuning.

% \autoref{fig:star-trial3-joint-errors} examines the most challenging of the four primary evaluation trajectories. At Trial~3, \DiffILC{} substantially suppresses the repetitive joint-space tracking errors observed with fixed-parameter model-based ILC. The joint RMSE decreases from $2.06^\circ$ to $0.52^\circ$, corresponding to a $74.8\%$ reduction relative to the fixed baseline, while the Cartesian RMSE decreases from $5.68~\mathrm{mm}$ to $1.82~\mathrm{mm}$, a $68.0\%$ reduction. The larger initial and residual errors for the star are consistent with its sharp changes in path direction. Nevertheless, the ten-epoch \DiffILC{} configuration achieves a substantial reduction within the first three physical repetitions without trajectory-specific retuning.

\begin{figure}[h]
    \centering
    \includegraphics[width=\linewidth]{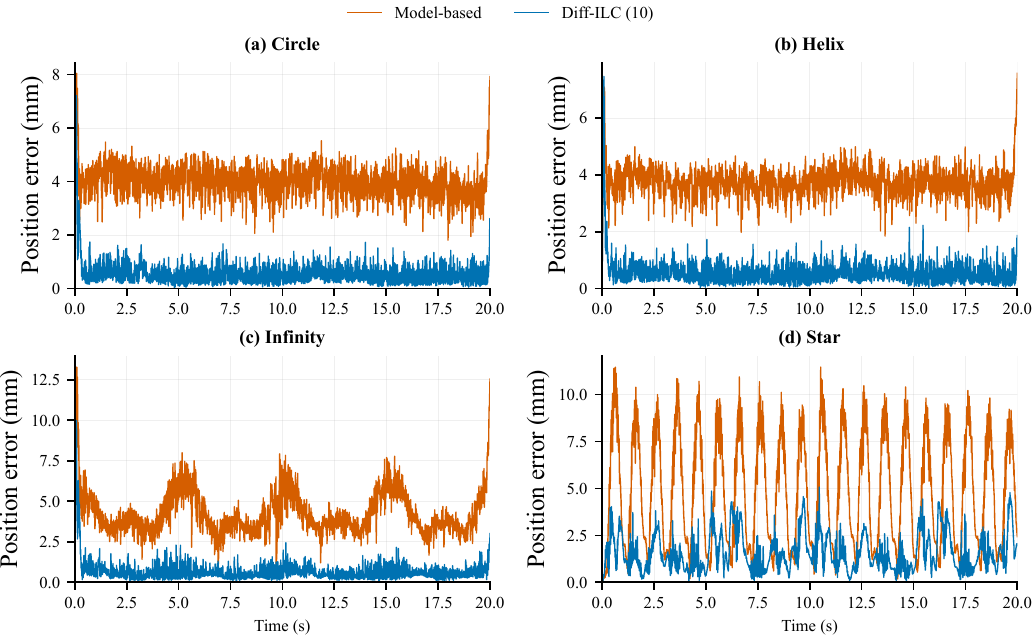}
    \caption{Cartesian position-error magnitude along the four primary evaluation trajectories at Trial~3 for fixed-parameter model-based ILC and \DiffILC{} with ten outer optimization epochs.}
    \label{fig:trial3-cartesian-errors}
    \vspace{-4mm}
\end{figure}

\autoref{fig:trial3-cartesian-errors} shows that, at Trial~3, \DiffILC{} reduces the Cartesian RMSE for each trajectory from $4.05$, $3.79$, $4.48$, and $5.68~\mathrm{mm}$ to $0.73$, $0.73$, $0.95$, and $1.82~\mathrm{mm}$, respectively. These results correspond to reductions of approximately $68\%$--$82\%$ relative to fixed-parameter model-based ILC across both smooth and high-curvature paths.

% \autoref{fig:trial3-cartesian-errors} shows the Cartesian position-error magnitude along each trajectory at Trial~3. \DiffILC{} reduces the Cartesian RMSE from $4.05$ to $0.73~\mathrm{mm}$ for the circle, from $3.79$ to $0.73~\mathrm{mm}$ for the helix, from $4.48$ to $0.95~\mathrm{mm}$ for the infinity trajectory, and from $5.68$ to $1.82~\mathrm{mm}$ for the star. These correspond to reductions of approximately $82.0\%$, $80.7\%$, $78.8\%$, and $68.0\%$, respectively, relative to fixed-parameter model-based ILC. The improvement is therefore observed across both smooth and high-curvature trajectory geometries rather than being confined to a single evaluation path.

The fast-circle experiment tested transfer to a higher execution speed using the same identified differential model and without speed-specific baseline retuning. At $0.15~\mathrm{Hz}$, \DiffILC{} reduced the Trial~3 joint/Cartesian RMSE from $2.40^\circ/6.02~\mathrm{mm}$ to $0.60^\circ/1.45~\mathrm{mm}$, corresponding to reductions of $75.0\%$ and $75.9\%$.  Thus, the early-trial advantage is retained for substantially faster motion despite reusing the same fixed differential model.

% The additional fast-circle experiment further evaluates whether this improvement persists away from the nominal execution condition. The $0.15~\mathrm{Hz}$ trajectory was executed using the same differential model identified once before the ILC experiments and without speed-specific retuning of the fixed baseline. At Trial~3, \DiffILC{} reduces the joint/Cartesian RMSE from $2.40^\circ/6.02~\mathrm{mm}$ to $0.60^\circ/1.45~\mathrm{mm}$, corresponding to reductions of $75.0\%$ and $75.9\%$, respectively. Thus, the early-trial advantage is retained for substantially faster motion despite reusing the same fixed differential model.
\vspace{-5mm}
\subsection{Online Parameter Adaptation}

\begin{figure}[t]
    \centering
    \includegraphics[width=\columnwidth]{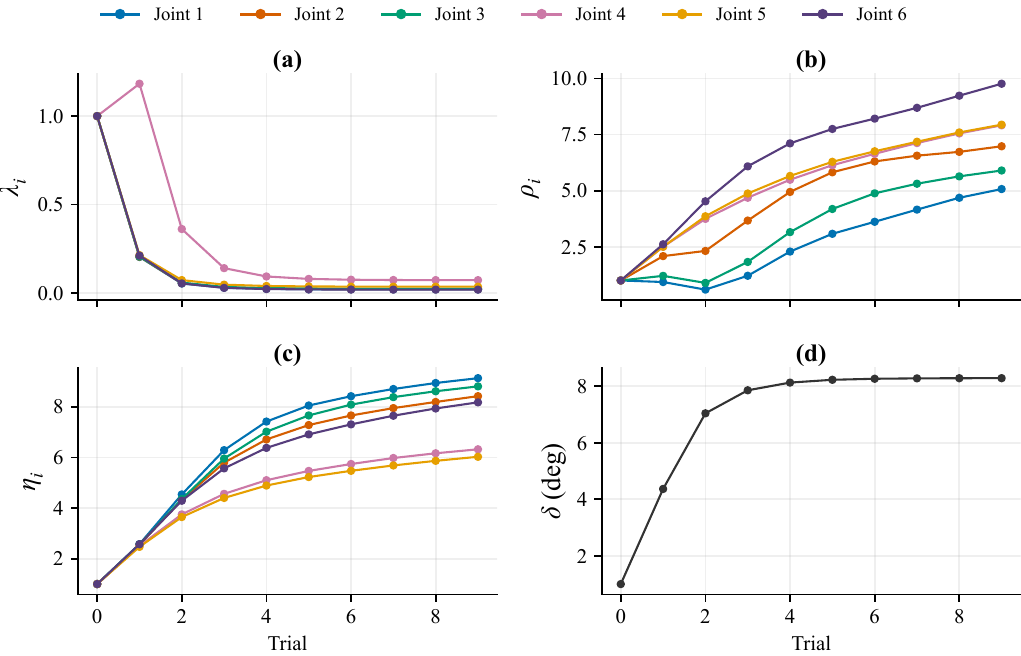}
    \caption{Online evolution of the learned \DiffILC{} (10) parameters for the star trajectory (a) correction penalties $\lambda_i$, (b) first-difference penalties $\rho_i$, (c) second-difference penalties $\eta_i$, and (d) shared correction bound $\delta$.}
    \label{fig:star-parameter-evolution}
    \vspace{-6mm}
\end{figure}

\autoref{fig:star-parameter-evolution} shows the online parameter evolution for the star trajectory. The correction penalties $\lambda_i$ decrease from $1$ to $0.019$--$0.073$, while $\rho_i$ and $\eta_i$ increase to $5.07$--$9.76$ and $6.03$--$9.13$, respectively; the correction bound $\delta$ increases from $1^\circ$ to $8.29^\circ$. Thus, \DiffILC{} permits larger corrections while increasing penalties on temporal variation and curvature, balancing correction magnitude and smoothness.
% Parameters are updated after Trials~1--9 using the latest measured response and applied to the subsequent trial.

% \autoref{fig:star-parameter-evolution} illustrates how the ILC design parameters evolve online for the star trajectory. The correction penalties $\lambda_i$ decrease from $1$ to $0.019$--$0.073$, while the first- and second-difference penalties $\rho_i$ and $\eta_i$ increase to $5.07$--$9.76$ and $6.03$--$9.13$, respectively. Meanwhile, the shared correction bound $\delta$ increases from $1^\circ$ to $8.29^\circ$. The learned parameter evolution therefore shifts the inner ILC problem toward larger-amplitude corrections through the decrease in $\lambda_i$ and relaxation of $\delta$, while simultaneously increasing the penalties on temporal variation and curvature through $\rho_i$ and $\eta_i$. The adaptation therefore does not simply make the ILC update uniformly more aggressive; instead, it jointly adjusts correction magnitude and temporal smoothness according to the task-level objective. Parameter adaptation is performed after Trials~1--9 using the latest measured response, and the resulting parameters are used to compute the correction applied on the subsequent physical trial.

Across the four unseen $0.1$~Hz trajectories, \DiffILC{} with ten outer epochs reduced the average Trial~3 joint and Cartesian RMSE by $78.8\%$ and $76.4\%$, respectively, relative to fixed-parameter model-based ILC. The consistent improvement across these trajectories and the faster-circle experiment indicates that the early-trial advantage extends across changes in path geometry and execution rate while using the same identified differential model.

% Overall, the experiments show that the primary advantage of \DiffILC{} is
% faster tracking-error reduction during the first few physical repetitions.
% Across the four unseen $0.1$~Hz trajectory geometries, the ten-epoch
% configuration reduces the average Trial~3 joint and Cartesian RMSE by
% $78.8\%$ and $76.4\%$, respectively, relative to fixed-parameter
% model-based ILC. The corresponding Trial~3 joint-RMSE reductions are
% $81.1\%$, $80.8\%$, $79.4\%$, and $74.8\%$ for the circle, helix,
% infinity, and star trajectories, respectively. The consistent early-trial
% advantage across these distinct trajectory geometries, together with the
% additional faster-motion experiment, provides evidence that the benefit is
% not specific to a single reference trajectory or execution condition.
% \DiffILC{} therefore improves physical-trial efficiency across changes in
% trajectory geometry and execution rate while reusing the same identified
% differential model and without trajectory- or speed-specific retuning of
% the fixed baseline.

% All applied command trajectories satisfied the prescribed joint-position,
% velocity, acceleration, and trial-to-trial correction constraints within
% the numerical feasibility tolerance of the QP solver.
\vspace{-5mm}
\section{Conclusion}
\label{sec:conclusion}

This work presented \DiffILC{}, a differentiable constrained model-based ILC framework for black-box position-commanded manipulators. A data-driven closed-loop differential model is combined with a finite-horizon constrained QP, while the ILC regularization weights and correction bound are adapted online through hypergradients obtained from the locally active KKT system. Experiments on a 6-DoF DOBOT CR3 showed consistent improvements across unseen planar and three-dimensional trajectories using a single identified differential model. With ten outer epochs, the mean joint/Cartesian RMSE at Trial~3 decreased to $0.35^\circ/1.06~\mathrm{mm}$, compared with $1.65^\circ/4.50~\mathrm{mm}$ for fixed-parameter model-based ILC. The results show that the principal benefit of \DiffILC{} is faster tracking-error reduction during the first few physical repetitions, while additional outer epochs yield diminishing returns.

A limitation of the current formulation is that parameter adaptation relies on a real-anchored next-trial surrogate constructed from a fixed identified differential model. While this enables several computational updates between robot executions, it can increase sensitivity to model mismatch. Future work will investigate physical-loss and data-driven hypergradient formulations. %that update the learning law and local command-to-motion sensitivity from subsequent measured robot performance.

\vspace{-5mm}
\bibliographystyle{IEEEtranN}
\bibliography{IEEEabrv,references}

\vfill

\end{document}